# Modelling transition dynamics in MDPs with RKHS embeddings

**Steffen Grünewälder**[1] STEFFEN@CS.UCL.AC.UK
**Guy Lever**[1] G.LEVER@CS.UCL.AC.UK
**Luca Baldassarre** L.BALDASSARRE@CS.UCL.AC.UK
**Massimilano Pontil** M.PONTIL@CS.UCL.AC.UK
**Arthur Gretton**[*] ARTHUR.GRETTON@GMAIL.COM
CSML and [*]Gatsby Unit, University College London, UK, [*] MPI for Intelligent Systems

## Abstract

We propose a new, nonparametric approach to learning and representing transition dynamics in Markov decision processes (MDPs), which can be combined easily with dynamic programming methods for policy optimisation and value estimation. This approach makes use of a recently developed representation of conditional distributions as *embeddings* in a reproducing kernel Hilbert space (RKHS). Such representations bypass the need for estimating transition probabilities or densities, and apply to any domain on which kernels can be defined. This avoids the need to calculate intractable integrals, since expectations are represented as RKHS inner products whose computation has linear complexity in the number of points used to represent the embedding. We provide guarantees for the proposed applications in MDPs: in the context of a value iteration algorithm, we prove convergence to either the optimal policy, or to the closest projection of the optimal policy in our model class (an RKHS), under reasonable assumptions. In experiments, we investigate a learning task in a typical classical control setting (the under-actuated pendulum), and on a navigation problem where only images from a sensor are observed. For policy optimisation we compare with least-squares policy iteration where a Gaussian process is used for value function estimation. For value estimation we also compare to the NPDP method. Our approach achieves better performance in all experiments.

## 1. Introduction

### 1.1. Preliminaries

Throughout we denote expectations by $\mathbb{E}[\cdot]$, and the probability over events by $\mathbb{P}(\cdot)$. We denote by $\mathcal{B}(\mathcal{X})$ and $\mathcal{C}_b(\mathcal{X})$ the Banach spaces of bounded functions and bounded continuous functions on $\mathcal{X}$, each equipped with the sup-norm $||\cdot||_\infty$.

We consider in particular the problem in which we control a trajectory $\{x_t\}_{t=0}^\infty$ over $\mathcal{X}$ by sequentially choosing actions $a_t \in \mathcal{A}$ at each time step $t \geq 0$, once $x_t$ is revealed, after which we receive a reward $r_{t+1} = r(x_t, a_t)$. We denote a set of deterministic *policies* $\Pi = \mathcal{A}^\mathcal{X}$. The objective is to find a policy $\pi$ which maximises the expected sum of rewards obtained by following $\pi$: $\mathbb{E}\left[\sum_{t=0}^\infty \gamma^t r_{t+1}(X_t, A_t)|X_0 = x, A_t = \pi(X_t)\right]$.

For a policy $\pi \in \Pi$ we denote the associated *value function*,

$$V^\pi(x) := \mathbb{E}\left[\sum_{t=0}^\infty \gamma^t r_{t+1}(X_t, A_t)|X_0 = x, A_t = \pi(X_t)\right],$$

and recall that $V^\pi(x) = r(x, \pi(x)) + \gamma \mathbb{E}_{X\sim P(\cdot|x,\pi(x))}[V^\pi(X)]$. We define the optimal value function $V^*(x) := \max_{\pi\in\Pi} V^\pi(x)$ for all $x \in \mathcal{X}$, and an optimal policy to be any $\pi^*$ such that $\pi^* \in \operatorname{argmax}_{\pi\in\Pi} V^\pi(x)$ for all $x \in \mathcal{X}$. For a given *action-value function* $Q : \mathcal{X} \times \mathcal{A} \to \mathbb{R}$ we define the *greedy policy* w.r.t. $Q$ by $\pi_Q(x) := \operatorname{argmax}_{a\in\mathcal{A}} Q(x, a)$ (choosing arbitrarily in the case of a tie) and the optimal action-value function,

$$Q^*(x, a) := r(x, a) + \gamma \mathbb{E}_{X\sim P(\cdot|x,a)}[V^*(X)], \qquad (1)$$

so that $\pi^* = \pi_{Q^*}$ (see e.g. (Szepesvari, 2009) for this background). We require the following well-known result, which is proved in the Appendix for reference (Grünewälder et al., 2012):

[1]**Equal contribution.**

**Lemma 1.1.** *(Singh & Yee, 1994)[Corollary 2] For any action-value function $Q : \mathcal{X} \times \mathcal{A} \to \mathbb{R}$, the greedy policy $\pi_Q$ satisfies $||V^{\pi_Q} - V^*||_\infty \leq \frac{2}{1-\gamma}||Q^* - Q||_\infty$.*

We are interested in the case where $P$ is unknown but a sample $\mathcal{S} := \{(x_i, a_i, x'_i)\}_{i=1}^m$ is provided, drawn i.i.d. from a distribution $\widetilde{P}$ such that $\widetilde{P}(X'_i = x'_i | X_i = x_i, A_i = a_i) = P(x'_i | x_i, a_i)$ for all $i$ (the marginal probabilities need not match). Note the abuse of notation here – subscripts index samples and not time steps.

### 1.2. Overview of the approach

A number of recent studies have focused on efficient evaluation of conditional expectations on functions that are "well behaved" in the sense that they belong to a reproducing kernel Hilbert space (RKHS). These approaches have been particularly successful in performing inference in graphical models, where the model parameters are learned nonparametrically from data (Song et al., 2010b; 2009; 2011). The key insight in these works is that conditional probabilities can be represented as functions in an RKHS, called *conditional distribution embeddings*. The conditional expectation of any function in the RKHS then becomes a linear operation, where we take the inner product with the appropriate distribution embedding.

Many methods for solving problems in MDPs require the computation of expectations of functions (value functions for example) with respect to transition dynamics, and so (approximations of) the operators

$$f \mapsto \mathbb{E}_{X \sim P(\cdot|x,a)}[f(X)] \tag{2}$$

are required. A direct but computationally costly approach would be to first learn a conditional density estimate (difficult in high dimensions), followed by (possibly intractable) integrals to compute the expectation. By contrast, our approach is a two stage process for learning in MDPs: we first use the theory of RKHS embeddings to estimate the operators (2) *directly* (over a specific class of functions in an RKHS), then use these estimated operators in standard approaches for solving MDPs – here we consider dynamic programming methods for value estimation and policy optimisation. The application to dynamic programming is described in more detail in Sec. 3.

### 1.3. Advantages of the approach

A direct kernel-based approach has a number of advantages. First, like density estimates, conditional embeddings can be learned from a training sample: we do not need to address the problem of modeling system dynamics, such as the differential equations governing a robot arm. Unlike density estimates, however, distribution embedding estimates do not scale poorly with the dimension $d$ of the underlying space: the risk of a kernel density estimate increases as $O(m^{-4/(4+d)})$ when the optimal bandwidth is used (Wasserman, 2006)[Sec. 6.5]. By contrast, the rate of convergence for conditional mean embeddings is independent of the dimension of the underlying space (Song et al., 2010b)[Thm. 1].

Second, the solution to many control problems involves computation of high dimensional integrals to obtain expectations, which is prohibitively costly. By contrast, RKHS embeddings explicitly provide a representation of the expectation operator as an RKHS inner product, which reduces calculating expectations to a computation of *linear complexity* in the number of training points used to represent the embedding, and avoids any intermediate problems such as density estimation and sample selection for numerical integration. Thus, the approach provides a framework for alleviating the curse of dimensionality in MDPs (particularly if, for example, sparsification of the embedding is considered, which we address briefly in the Appendix (Grünewälder et al., 2012)). The conditional distribution embeddings themselves may be computed exactly at cost cubic in the training sample size, and approximated to good accuracy at linear cost.

A third advantage is that we can provide convergence results in the infinite sample case. Thm. 3.2 demonstrates how a performance guarantee for value iteration using embeddings decomposes into guarantees for value iteration and gurantees for the embeddings, upper bounding the difference $||V^{\widehat{\pi}_\kappa} - V^*||_\infty$ between the optimal value $V^*$ and the value $V^{\widehat{\pi}_\kappa}$ of the policy $\widehat{\pi}_\kappa$ found by performing value iteration using the embeddings after $\kappa$ iterations. This bound contains a term involving how well we can approximate $V^*$ in our model class (a chosen RKHS) – which usually corresponds to smoothness assumptions on $V^*$ – and can decrease by increasing the richness of the RKHS. A second term captures how quickly we can learn the embeddings for the operator (2) over functions in the chosen RKHS. This bound can be specialised to give convergence guarantees for specific settings by plugging in guarantees for the two components: in Corollary 3.3, we specialise to the common setting of finite state space and positive definite kernel and obtain that $||V^{\widehat{\pi}_\kappa} - V^*||_\infty \to 0$.

As a final advantage, the method applies wherever kernels may be defined, including on high dimensional or continuous state spaces, manifolds (kernels on the surface of a sphere (Wendland, 2005) are of particu-lar interest in robotics), and partially observable tasks where only sensor measurements are available.

### 1.4. Relation to existing methods

Kernel methods have become increasingly popular in RL. Methods include kernel LSTD (Xu et al., 2005) and GPTD (Engel et al., 2005). Both can be used to estimate (q-)values and they differ in this mainly through the regulariser (Taylor & Parr, 2009). Based on the q-value estimates it is possible to optimise the policy. Other related approaches include Rasmussen & Kuss (2003), Deisenroth et al. (2009), Ormoneit & Sen (1999) and Kroemer & Peters (2011). Here, transition models (densities) are learned with the help of Gaussian processes or kernel density estimates. Using them for value estimation or policy optimisation usually leads to difficult integration to be solved numerically via, e.g., an intermediate sampling method. In contrast we use kernels to directly learn the expectation operators and avoid numerical integration. Finally, in (Parr et al., 2008) a way is proposed to approximate expectations in a low dimensional state representation. In contrast to our approach the paper assumes that the true expectation is known.

## 2. RKHS embeddings of transition probability kernels

Given a set $\mathcal{Z}$ and a positive semi-definite (p.s.d.) kernel $K : \mathcal{Z} \times \mathcal{Z} \to \mathbb{R}$ (see e.g. Steinwart & Christmann, 2008, for details) we denote by $\mathcal{H}_K \subseteq \mathbb{R}^{\mathcal{Z}}$ its unique reproducing kernel Hilbert space (RHKS), and by $\langle \cdot, \cdot \rangle_K$ the inner product in $\mathcal{H}_K$. Due to the reproducing property of $K$ in $\mathcal{H}_K$ we have $h(x) = \langle K(x, \cdot), h\rangle_K$ for all $h \in \mathcal{H}_K$. We recall the notion of a universal kernel: given a Banach space of functions $\mathcal{F} \subseteq \mathbb{R}^{\mathcal{Z}}$ a kernel is $\mathcal{F}$*-universal* if $\mathcal{H}_K$ is dense in $\mathcal{F}$. We denote $\rho_K := \sup_{z\in\mathcal{Z}} \sqrt{K(z, z)}$ and refer to kernels $K$ such that $\rho_K < \infty$ as bounded kernels.

Following Sriperumbudur et al. (2010), given any probability distribution $P$ and p.s.d. kernel $K$ on a set $\mathcal{Z}$ a *distribution embedding* of $P$ in $\mathcal{H}_K$ is an element $\mu \in \mathcal{H}_K$ such that $\langle \mu, h\rangle_K = \mathbb{E}_{Z\sim P}[h(Z)]$ for all $h \in \mathcal{H}_K$. In our application, given p.s.d. kernels $L : \mathcal{X}\times\mathcal{X} \to \mathbb{R}$ and $K : (\mathcal{X}\times\mathcal{A})\times(\mathcal{X}\times\mathcal{A}) \to \mathbb{R}$, we are interested in the embedding of the expectation operator (2) corresponding to the state transition probability kernel $P$, over the domain $\mathcal{H}_L$; that is, an element $\mu_{(x,a)} \in \mathcal{H}_L$ such that $\langle \mu_{(x,a)}, f\rangle_L = \mathbb{E}[f(X_{t+1})|X_t = x, A_t = a]$, for all $f \in \mathcal{H}_L$ and for all $t \geq 0$ – recall that the Markov property implies such a $\mu_{(x,a)}$ is independent of time. Recalling Sec. 1.1, given the sample $\mathcal{S}$, we will consider a sample-based estimate of the expectation operator (2). This will be achieved by identifying an element $\overline{\mu}_{(x,a)} \in \mathcal{H}_L$ such that, for all $f \in \mathcal{H}_L$, $\langle\overline{\mu}_{(x,a)}, f\rangle_L$ approximates $\mathbb{E}_{X\sim P(\cdot|x,a)}[f(X)]$. Following (Song et al., 2009; 2010b) an estimate is

$$\overline{\mu}_{(x,a)} := \sum_{i=1}^{m} \alpha_i(x, a)L(x'_i, \cdot) \in \mathcal{H}_L, \tag{3}$$

where $\alpha_i(x, a) = \sum_{j=1}^m W_{ij}K((x_j, a_j), (x, a))$, and where $\boldsymbol{W} := (\boldsymbol{K} + \lambda m\boldsymbol{I})^{-1}$, $\boldsymbol{K} = (K((x_i, a_i), (x_j, a_j)))_{ij=1}^m$, and $\lambda$ is a regularization parameter. We assume w.l.o.g. $x'_i \neq x'_j$ for all $x'_i, x'_j$ in the expansion (3).[1] In some situations, the estimate (3) is consistent in the RHKS norm sense and uniformly over $\mathcal{X}\times\mathcal{A}$: the following result, proved in the appendix, follows directly from (Song et al., 2010b)[Thm. 1].

**Lemma 2.1.** *Suppose $K$ is a bounded kernel and the conditions of (Song et al., 2010b)[Thm. 1] are satisfied.*[2] *Then* $\sup_{(x,a)\in\mathcal{X}\times\mathcal{A}}\{||\mu_{(x,a)} - \overline{\mu}_{(x,a)}||_L\} \in \mathcal{O}_{\widetilde{P}}(\lambda^{\frac{1}{2}} + \lambda^{-\frac{3}{2}}m^{-\frac{1}{2}})$, *and thus by choosing $\lambda \to 0$, $\lambda^3 m \to \infty$ we have that, for any $\epsilon > 0$,*

$$\mathbb{P}_{\mathcal{S}\sim\widetilde{P}^m}\left(\sup_{(x,a)\in\mathcal{X}\times\mathcal{A}} ||\mu_{(x,a)} - \overline{\mu}_{(x,a)}||_L > \epsilon\right) \to 0.$$

By the reproducing property of $L$, we have

$$\langle\overline{\mu}_{(x,a)}, f\rangle_L = \sum_{i=1}^{m} \alpha_i(x, a)f(x'_i)$$

In this work, for theoretical analysis, we consider a normalised version of (3):

$$\widehat{\mu}_{(x,a)} := \sum_{i=1}^{m} \widehat{\alpha}_i(x, a)L(x'_i, \cdot) \in \mathcal{H}_L, \tag{4}$$

where $\widehat{\alpha}_i(x, a) = \frac{\alpha_i(x,a)}{\sum_{j=1}^m |\alpha_j(x,a)|}$. This is a technical consideration which will later ensure that we can define a

[1] We can otherwise form a new expansion in which the $x'_i$ are unique by summing any $\alpha_i(x, a)$ as necessary.

[2] These conditions require that the mapping $(x, a) \mapsto \mathbb{E}_{X\sim P(\cdot|(x,a))}[f(X)]$ be an element of $\mathcal{H}_K$ for all $f \in \mathcal{H}_L$, and that the operator $C_{YX}C_{XX}^{-3/2}$ be Hilbert-Schmidt, where $C_{YX}$ and $C_{XX}$ are covariance operators: see (Song et al., 2009) or Appendix D.1 for details (Grünewälder et al., 2012). The first condition is a smoothness assumption on the distribution, and for the convergence guarantee of Corollary 3.3 we specialise to the simple setting of finite state space, in which case this condition is trivially satisfied.The second condition is guaranteed in our case when, for example, the marginal density of the initial state $X$ from $\widetilde{P}$ is bounded away from zero and the RKHSs $\mathcal{H}_K$, $\mathcal{H}_L$ are of finite dimensionality.

**Algorithm 1 Estimate Conditional Expectation**

**input** Sample of transitions $\mathcal{S} := \{(x_1, a_1, x_1'), \ldots, (x_m, a_m, x_m')\}$, kernel $K$ on $\mathcal{X} \times \mathcal{A}$ and a kernel $L$ on states $\mathcal{X}$
**output** A conditional expectation estimate $\overline{\mu}_{(x,a)}$
Build kernel matrix $\boldsymbol{K}$ for samples $\{(x_1, a_1), \ldots, (x_m, a_m)\}$
Calculate coefficient vector $\alpha_i := \sum_{j \leq m} \boldsymbol{W}_{ij} K((x_j, a_j), (x, a))$, where $\boldsymbol{W} := (\boldsymbol{K} + \lambda m \boldsymbol{I})^{-1}$
Calculate the estimate $\overline{\mu}_{(x,a)} := \sum_{i=1}^m \alpha_i(x, a) L(x_i', \cdot)$

**Algorithm 2 Estimate Value**

**input** Sample $\mathcal{S}$, policy $\pi$, conditional expectation estimate $\overline{\mu}_{(x,a)}$, discount $\gamma$, max. number of iterations $N$, error threshold $\theta$, reward function $r$
**output** Value estimate $\hat{V}$
$n = 1, error = 1$; define a value vector for states $x_1', ..., x_m'$: $V := \mathbf{0}$
**while** $n \leq N$ **and** $error > \theta$ **do**
  **for all** $i \leq m$ **do**
    $V'(x_i') \leftarrow r(x_i', \pi(x_i')) + \gamma \langle \overline{\mu}_{(x_i', \pi(x_i'))}, V \rangle_L$
  **end for**
  $n \leftarrow n + 1$, $error \leftarrow ||V' - V||_\infty$, $V \leftarrow V'$
**end while**
**return** $\hat{V}(x) = r(x, \pi(x)) + \gamma \langle \overline{\mu}_{(x,\pi(x))}, V \rangle_L$

**Algorithm 3 Approximate Value Iteration**

**input** Sample $\mathcal{S}$, discount $\gamma$, maximum number of iterations $N$, reward function $r$, error threshold $\theta$
**output** $\overline{\mu}_{(x,a)}$, approximate optimal value $\hat{V}$
$n = 1, error = 1$; define a value vector for states $x_1', ..., x_m'$: $V := \mathbf{0}$
Run Alg. 1 to get $\overline{\mu}_{(x,a)}$
**while** $n \leq N$ **and** $error > \theta$ **do**
  **for all** $i \leq m$ **do**
    $V'(x_i') \leftarrow \max_{a \in \mathcal{A}} r(x_i', a) + \gamma \langle \overline{\mu}_{(x_i', a)}, V \rangle_L$
  **end for**
  $n \leftarrow n + 1$, $error \leftarrow ||V' - V||_\infty$, $V \leftarrow V'$
**end while**
**return** $\hat{V}(x) = \max_{a \in \mathcal{A}} r(x, a) + \gamma \langle \overline{\mu}_{(x,a)}, V \rangle_L$

certain contraction mapping. We now demonstrate the consistency of the estimators defined by (4) for finite state spaces, by showing that in the limit of large data the normalization of $\widehat{\alpha}$ has no effect. The following lemma is proved in the Appendix (Grünewälder et al., 2012).

**Lemma 2.2.** *Under the conditions of Lemma 2.1, and if $|\mathcal{X}| < \infty$ and $L$ is strictly positive definite, by choosing $\lambda \to 0$, $\lambda^3 m \to \infty$ we have that, for any $\epsilon > 0$,*

$$\mathbb{P}_{\mathcal{S} \sim \widetilde{P}^m} \left( \sup_{(x,a) \in \mathcal{X} \times \mathcal{A}} ||\mu_{(x,a)} - \widehat{\mu}_{(x,a)}||_L > \epsilon \right) \to 0.$$

## 3. Application to MDPs

The learnt embeddings are applied to MDPs by recalling (4) and defining an operator

$$\widehat{\mathscr{E}}_{(x,a)}[f] := \sum_{i=1}^{m} \widehat{\alpha}_i(x, a) f(x_i'). \tag{5}$$

When $f \in \mathcal{H}_L$ we have that $\widehat{\mathscr{E}}_{(x,a)}[f] = \langle \widehat{\mu}_{(x,a)}, f \rangle_L \approx \mathbb{E}_{X \sim P(\cdot|(x,a))}[f(x)]$. When $f \notin \mathcal{H}_L$ the quality of the approximation will further depend upon how well $f$ can be approximated by a low norm function in $\mathcal{H}_L$. This operator can be used in place of the true unknown expectation operator (2) in any MDP method which makes use of such expectations, such as dynamic programming. As an example below, we analyse value iteration, but similar considerations yield similar analyses for other methods. We summarize a joint value estimation algorithm and policy optimisation approach in the Algorithm boxes above.

If we knew $P$, and could efficiently compute expectations, we could define the Bellman operator $B$ as

$$(BV)(x) := \max_{a \in \mathcal{A}} \{r(x, a) + \gamma \mathbb{E}_{X \sim P(\cdot|x,a)}[V(X)]\}, \tag{6}$$

where we suppose that the image of $B$ is always a measurable function.[3] Recall that picking an arbitrary $V_0$ and iterating $V_{k+1} = BV_k$ converges in sup-norm, $V_k \to V^*$ (see e.g. Szepesvari, 2009). Since we do not know $P$, we use the embeddings $\widehat{\mu}_{(x,a)}$ and, recalling (5), define the operator $\widehat{B} : \mathcal{B}(\mathcal{X}) \to \mathcal{B}(\mathcal{X})$ as

$$(\widehat{B}V)(x) := \max_{a \in \mathcal{A}} \{r(x, a) + \gamma \widehat{\mathscr{E}}_{(x,a)}[V]\}. \tag{7}$$

It is necessary to define $\widehat{B}$ on functions which are not in $\mathcal{H}_L$, and this possibility introduces a term in the analysis which captures how well $V^*$ can be approximated in $\mathcal{H}_L$ (See Thm. 3.2). By Lemma 2.2, in the limit of large data, the operator defined by (7) converges to an expectation operator on functions in $\mathcal{H}_L$, and thus $\widehat{B}$ can be seen to approximate $B$ defined by (6) on $\mathcal{H}_L$. The following result is proved in the Appendix (Grünewälder et al., 2012):

**Proposition 3.1.** *$\widehat{B}$ is a sup-norm contraction on the space $\mathcal{B}(\mathcal{X})$ with Lipschitz constant $\gamma$.*

Since $\widehat{B}$ defines a sup-norm contraction mapping on a complete metric space, by Banach's fixed point theorem (e.g. Granas & Dugundji, 2003) there exists a unique fixed point $\widehat{V}^*$ of $\widehat{B}$, such that choosing $\widehat{V}_0$ arbitrarily and iterating $\widehat{V}_{k+1} = \widehat{B}\widehat{V}_k$ converges, $\widehat{V}_k \to \widehat{V}^*$, in sup-norm,

$$||\widehat{V}_k - \widehat{V}^*||_\infty \leq \frac{\gamma^k}{1-\gamma}||\widehat{V}_1 - \widehat{V}_0||_\infty. \tag{8}$$

[3]We suppose for simplicity that any necessary conditions to ensure this are met, since strictly speaking $B$ is defined only on measurable functions, see for example (Bertsekas & Shreve, 1978) for a discussion of the issues. In particular, these conditions are met when $|\mathcal{X}| < \infty$.

Suppose we perform $\kappa$ iterations, obtaining the estimate $\widehat{V}_\kappa \approx \widehat{V}^*$. Once $\widehat{V}_\kappa$ is obtained we form a policy $\widehat{\pi}_\kappa$ on-the-fly[4] by acting greedily w.r.t. $\widehat{Q}_\kappa(x,a)$, where

$$\widehat{Q}_\kappa(x,a) := r(x,a) + \gamma\widehat{\mathscr{E}}_{(x,a)}[\widehat{V}_\kappa], \tag{9}$$

so that the learned policy is

$$\widehat{\pi}_\kappa(x) := \operatorname*{argmax}_{a\in\mathcal{A}} \widehat{Q}_\kappa(x,a). \tag{10}$$

for each $x$ in a trajectory.

**Consistency:** We now discuss the consistency of $\widehat{\pi}_\kappa$ as an estimate of an optimal policy $\pi^*$. The following theorem decomposes the convergence of $V^{\widehat{\pi}_\kappa}$ to the optimal value function $V^*$, in terms of the convergence of value iteration, the convergence of the embeddings and how well we can approximate $V^*$ in sup-norm by a (low $||\cdot||_L$-norm) function in $\mathcal{H}_L$. This is a generic bound into which we can plug any suitable guarntees for an embedding method. In Corollary 3.3, we specialise the result to the finite state space case, where we can approximate $V^*$ arbitrarily well.

**Theorem 3.2.**

$$||V^{\widehat{\pi}_\kappa} - V^*||_\infty \leq \frac{2\gamma}{(1-\gamma)^2}\Big(\gamma^\kappa||\widehat{V}_1 - \widehat{V}_0||_\infty + 2||V^* - \widetilde{V}^*||_\infty + \sup_{(x,a)}||\mu_{(x,a)} - \widehat{\mu}_{(x,a)}||_L||\widetilde{V}^*||_L\Big), \tag{11}$$

*where $\widetilde{V}^*$ is any element of $\mathcal{H}_L$. Thus, whenever $\sup_{(x,a)}||\mu_{(x,a)} - \widehat{\mu}_{(x,a)}||_L \to 0$ in $\widetilde{P}$-probability, we have that, for any chosen $\widetilde{V}^* \in \mathcal{H}_L$,*

$$||V^{\widehat{\pi}_\kappa} - V^*||_\infty \leq \frac{4\gamma}{(1-\gamma)^2}||V^* - \widetilde{V}^*||_\infty + \epsilon_\kappa + \epsilon_m, \tag{12}$$

*where $\epsilon_\kappa \to 0$ and $\epsilon_m \to 0$ with convergence in $\widetilde{P}$-probability.*

*Proof.* (Sketch, see Appendix for full proof (Grünewälder et al., 2012).) The proof hinges upon obtaining the following chain of convergences,

$$\widehat{\mathscr{E}}_{(x,a)}[\widehat{V}_\kappa] \to_{(a)} \widehat{\mathscr{E}}_{(x,a)}[\widehat{V}^*] \approx_{(b)} \widehat{\mathscr{E}}_{(x,a)}[V^*] \approx \widehat{\mathscr{E}}_{(x,a)}[\widetilde{V}^*] \to_{(c)} \mathbb{E}_{X\sim P(\cdot|(x,a))}[\widetilde{V}^*(X)] \approx \mathbb{E}_{X\sim P(\cdot|(x,a))}[V^*(X)].$$

The convergence $(a)$ is a standard result for contraction mappings, $(b)$ requires a new lemma relating the fixed points of similar contraction mappings, and $(c)$ is possible using Lemma 2.2 because $\widetilde{V}^* \in \mathcal{H}_L$. Once this is obtained we recall that $\widehat{\pi}_\kappa$ is greedy w.r.t. $\widehat{Q}_\kappa$ defined by (9), and apply Lemma 1.1, since the optimal policy is greedy w.r.t. $Q^*$. □

We now interpret Thm. 3.2. The upper bound is,

$$||V^{\widehat{\pi}_\kappa} - V^*||_\infty \leq \frac{2\gamma}{(1-\gamma)^2}\Big(\overbrace{\gamma^\kappa||\widehat{V}_1 - \widehat{V}_0||_\infty}^{(i)} + \underbrace{2||V^* - \widetilde{V}^*||_\infty}_{(ii)} + \underbrace{\sup_{(x,a)}||\mu_{(x,a)} - \widehat{\mu}_{(x,a)}||_L||\widetilde{V}^*||_L}_{(iii)}\Big).$$

Here (i) is the standard difference between the value estimate of the initial policy and the value estimate of the policy that we get after applying one dynamic programming update. This term decreases to 0 with growing $\kappa$ because $\gamma < 1$. (ii) is the distance from the optimal value $V^*$ to any approximation $\widetilde{V}^*$ in the RKHS, and is therefore small when $\widetilde{V}^*$ is close to $V^*$ and so can be smaller when $\mathcal{H}_L$ is chosen to be a richer class. Finally, (iii) measures the quality of the learned embedding: $||\mu_{(x,a)} - \widehat{\mu}_{(x,a)}||_L$ is the distance between the empirical estimate $\widehat{\mu}$ of the conditional distribution embedding of $x'$ given $(x,a)$, and the population conditional embedding $\mu$, measured in the RKHS with kernel $L$. This difference is weighted by $||\widetilde{V}^*||_L$, the RKHS norm of the approximation $\widetilde{V}^*$. Intuitively, a lower RKHS norm implies a smoother function: when the norm is smaller, $\widetilde{V}^*$ is smoother, and the convergence faster. Thus (iii) requires us to obtain a better conditional mean embedding (via more training samples) when the value function is non-smooth. In other words, our approach favors smooth value functions, although given sufficient evidence, non-smooth functions can also be learned. One specialization is to the case when $V^* \in \mathcal{C}_b(\mathcal{X})$ and $L$ is a $\mathcal{C}_b(\mathcal{X})$-universal kernel (Steinwart & Christmann, 2008, Section 4.6). In this case we can choose $\widetilde{V}^*$ such that $||V^* - \widetilde{V}^*||_\infty$ is arbitrarily small in (11).

We now specialise Thm. 3.2 to the case where $|\mathcal{X}| < \infty$ and where $L$ is strictly positive definite kernel on $\mathcal{X}$ (we then know from Lemma 2.2 that $\sup_{(x,a)}||\mu_{(x,a)} - \widehat{\mu}_{(x,a)}||_L \to 0$ and that all real-valued functions are in the associated RKHS). Thus consistency is attained in otherwise very general conditions – the following is proved in the appendix:

**Corollary 3.3.** *Let $|\mathcal{X}| < \infty$ and $L$ be strictly positive definite. Under the conditions of Lemma 2.2 we*

[4] Meaning that we only need to calculate $\widehat{\pi}_\kappa(x)$ at points $x$ in a trajectory as and when required.

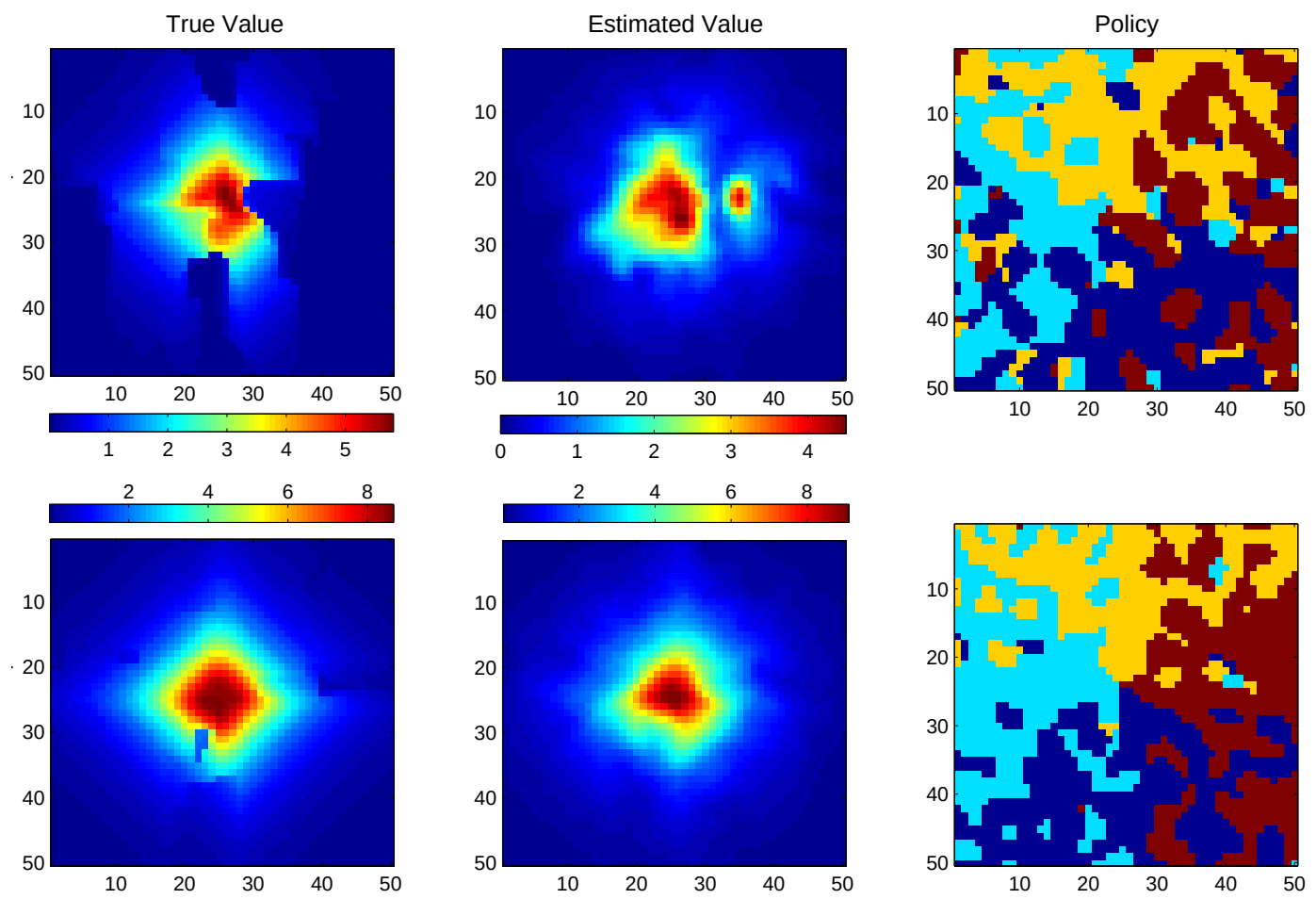

*Figure 1.* The left column shows the (true) value of the learned policy (color coded). The middle column shows the estimated value and the right column shows the policy. Actions are shown in the plot via a color code: yellow: go down; brown: left; dark blue: up; light blue: right. The 5000 sample policy is better (see for example the scale on the value color bars) and estimated value is close to true value. The patchy coloring is not a problem as, for example, in the bottom right it does not matter if the agent first goes up or to the left. The method has essentially learnt the task.

*have that* $||V^{\widehat{\pi}_\kappa} - V^*||_\infty \to 0$ *with convergence in* $\widetilde{P}$-*probability.*

**Complexity analysis:** Once the embeddings are learnt, the complexity of learning the approximate value function $\widehat{Q}_\kappa$ is $\mathcal{O}(m^2|\mathcal{A}|\kappa)$: due to the expansion of $\widehat{\mu}_{(x,a)}$ in the $m$ points in $\mathcal{S}$, computing each expectation is $\mathcal{O}(m)$ and we only ever need to know the evaluation of each iterate $\widehat{V}_k$ at the $m$ points in $\mathcal{S}$. Applying the learnt policy (10) to a trajectory $(x_0, x_1, \dots, x_T)$ of length $T$, is similarly $\mathcal{O}(m|\mathcal{A}|T)$. In Sec. B of the Supplementary material, we propose a sparser representation of the embedding, using an incomplete Cholesky approximation (Shawe-Taylor & Cristianini, 2004)[Sec. 5.2]. This reduces the cost of learning the embeddings from cubic to linear in $m$, and allows us to compute subsequent expectations in $\mathcal{O}(\ell)$, where generally $\ell \ll m$.

# 4. Experiments

We performed three experiments, using the embeddings in value estimation and policy optimization. The first experiment was an MDP with a fully observed discrete state space, to demonstrate convergence of the value function with increasing training sample size. The second and third experiments evaluate our approach on a classical control task and a task with high dimensional states. In policy optimisation we compare to LSPI (Lagoudakis & Parr, 2003) where we use the q-value estimator from (Engel et al., 2005), and for value estimation we compare to NPDP[5] (Kroemer & Peters, 2011). We achieve better performance in all our experiments.

We briefly address the choice of the regularization term $\lambda$. It can be shown that the conditional embeddings solve,

$$\hat{\mu} := \operatorname*{argmin}_{\mu \in \mathcal{H}} \left[ \sum_{i=1}^{m} \|L(x_i', \cdot) - \mu(x_i, a_i)\|_L^2 + \lambda \|\mu\|_{\mathcal{H}}^2 \right].$$

where $\mathcal{H} \subseteq (\mathcal{H}_L)^{(\mathcal{X}\times\mathcal{A})}$, recovering the vector-valued regression setting of Micchelli & Pontil (2005) (see Sec. D for details) which provides cross validation scheme for the parameter $\lambda$.

## 4.1. Experiment 1

The first experiment is a navigation experiment in a 50 x 50 room. The reward is a Gaussian centered in the middle of the room. The agent has four actions: go north, east, south or west. Each action has a success rate of 80 % and results in random movement with 20 % chance. The state space is fully observed. We learn the conditional distribution embedding from either 1000 or 5000 uniformly sampled transitions, uniformity ensuring we avoid exploration artifacts. We used a Gaussian kernel and cross-validated to determine the regulariser. Results are shown in Figure 1.

## 4.2. Experiment 2

We consider the under-actuated pendulum swing up task (Deisenroth et al., 2009). We generate a discrete-time approximation of the continuous-time pendulum dynamics as done in (Deisenroth et al., 2009). Starting from an arbitrary state the goal is to swing the pendulum up and balance it in the inverted position. The applied torque is $u \in [-5, 5]Nm$ and is not sufficient for a direct swing up. The state space is defined by the angle $\theta \in [-\pi, \pi]$ and the angular velocity, $\omega \in [-7, 7]$. The reward is given by the function $r(\theta, \omega) = \exp(-\theta^2 - 0.2\omega^2)$. For policy learning we compared to the GP-based LSPI approach and for value learning to NPDP. The results of the comparison are shown in Fig. 2.

[5]We thank the authors for providing code.

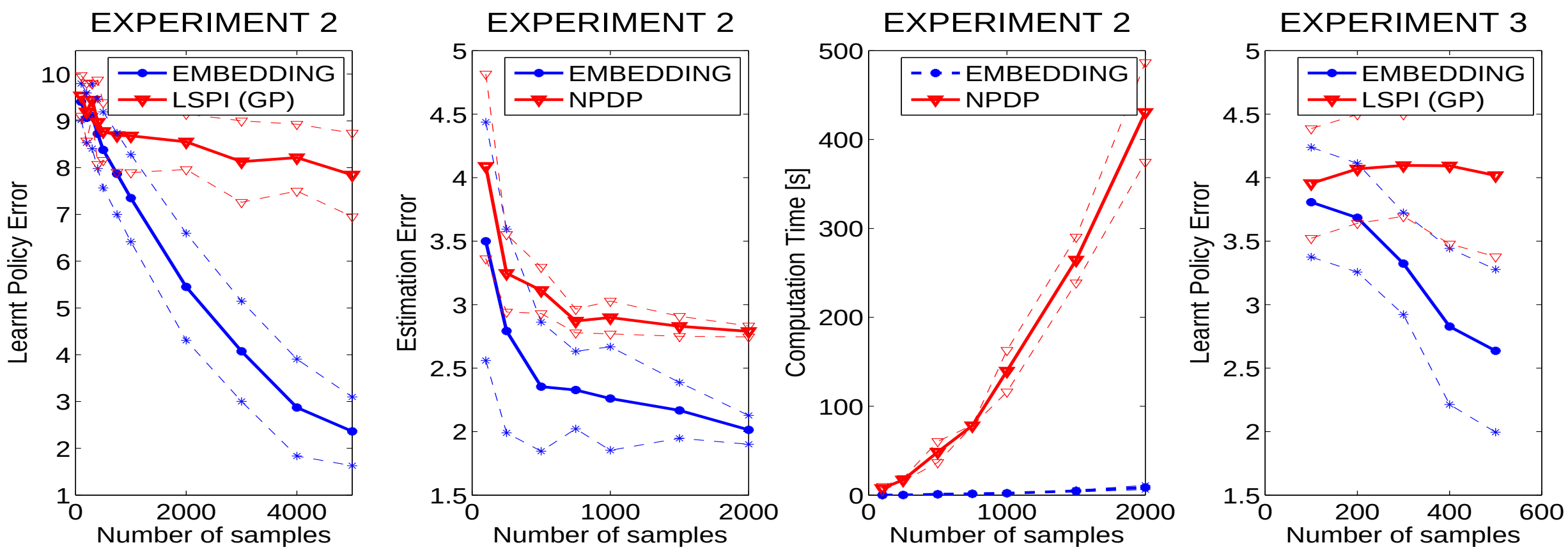

*Figure 2.* We compared our approach in policy learning to the GP based LSPI approach (1st and 4th plot) and in value learning to NPDP (2nd and 3rd plot). 1st and 4th plot: y-axis shows the average difference of the value of the learnt policy to the optimal value (averaged over the state space). In experiment 2 LSPI improves slowly after 500 samples and has problems with the two room task for small samples. 2nd plot: y-axis shows the prediction error of the value estimators (averaged over the state space). The embedding-based estimator is significantly better, especially for higher sample numbers. 3rd plot: run time for the two methods – the embedding method is 50-110 times faster on this task.

**Details for the policy learning setting:** We sampled uniformly from the state and action space and used a Gaussian kernel on both, selecting as kernel width the average K-neighbour distance, where K is one quarter of the sample size. We considered a discretization of the action space into 25 actions and we measured the difference between the value function evaluated on a grid of $25 \times 25$ points to the optimal value obtained by dynamic programming using the deterministic system dynamics. We compared over different sample sizes and averaged the performance over 10 repetitions.

**Details for the value estimation setting:** We used the optimal policy to generate samples. The goal was to predict the value of the optimal policy. The performance of NPDP depends strongly on the bandwidth parameter of the used kernel (a Gaussian). For parameter selection, we optimised performance on a validation set over a grid all free parameters (bandwidth for NPDP, bandwidth and $\lambda$ for the embedding), and report the error on an independent test set. The relatively poor scaling of NPDP with increased sample size is due to the numerical integration step in (Kroemer & Peters, 2011, Algorithm 1).

### 4.3. Experiment 3

Our final experiment is a high dimensional task where sensor measurements are available, and no state description is present. The environment consists of two rooms connected via a short corridor (Böhmer, 2012). The sensor measurements are images from a 3D renderer, and we aggregate four orientations (north, east, south and west) for a panorama, since the camera im-ages are ambiguous, especially close to the walls. The task of the agent is to reach a goal located in one of the rooms, using only the images to orient itself. Training points were chosen uniformly over the input space. We used a Gaussian kernel and cross-validated the regularization parameter. Results for 4000 training points are shown in Figure 3. We compared to the GP based LSPI approach using the same kernel and settings for both approaches; results are shown in Figure 2. Our method improves with increasing sample numbers. The GP based LSPI approach has obvious difficulties with this task and does not improve. We did not apply NPDP to exp. 3, as it would be computationally intractable in given the high dimensionality.

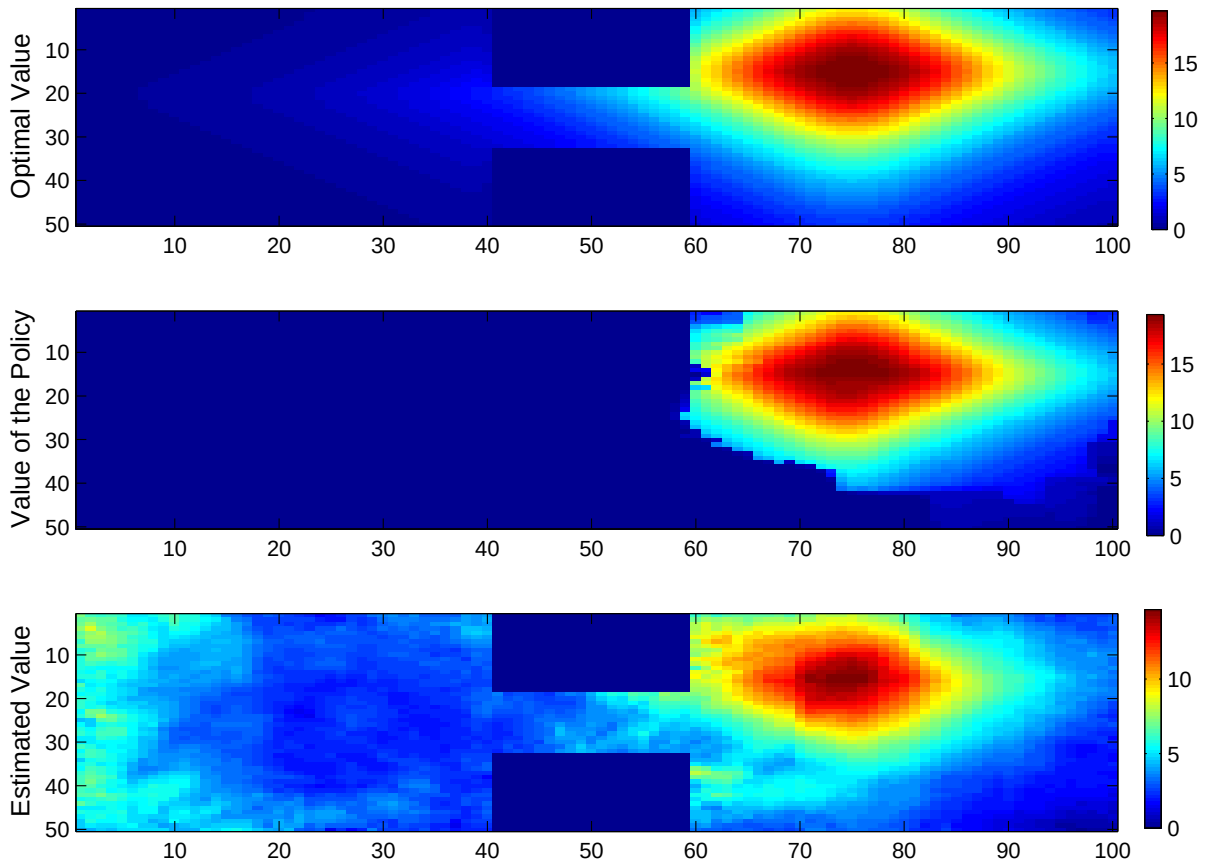

*Figure 3.* Results for Experiment 3 with 4000 data points. Top figure: optimal value. Middle figure: value of the learned policy. Bottom figure: the predicted value. The policy is nearly perfect in the room containing the objective. The performance degrades when the corridor is reached due to the challenging ambiguous nature of the images, which are insufficient to accurately distinguish between the locations. Similarly, the left wall has a high predicted value. The bottom picture shows that the value estimate is close to the optimal value.

## 5. Conclusions and Outlook

We have proposed a novel application of RKHS embeddings to learning expectation operators associated to transition dynamics in MDPs, with particular focus on their use in dynamic programming methods. The approach avoids the need for density estimates, sampling methods for evaluation of integrals, or explicit models of the system; is computationally efficient, having cost linear in the number of samples used in training (or even sublinear, with appropriate approximations); and has performance guarantees. Future work will focus on generalizing to more complex state and action spaces, and extending the convergence results to continuous state spaces. Another important generalization concerns the sampling distribution, which here is assumed to be iid, but one can expect similar results to hold in the non-iid case.

## Acknowledgements

The authors want to thank for the support of the EPSRC #EP/H017402/1 (CARDyAL) and the European Union #FP7-ICT-270327 (Complacs).